\documentclass[conference]{IEEEtran}

\usepackage{orcidlink}
\usepackage{cite}
\usepackage{amsmath,amssymb,amsfonts}
\usepackage{graphicx}
\usepackage{textcomp}
\usepackage{xcolor}
\def\BibTeX{{\rm B\kern-.05em{\sc i\kern-.025em b}\kern-.08em
    T\kern-.1667em\lower.7ex\hbox{E}\kern-.125emX}}

\usepackage{latexsym}

\usepackage{booktabs}
\usepackage{tabularx}
\newcolumntype{C}{>{\centering\arraybackslash}X}
\newcolumntype{R}{>{\raggedleft\arraybackslash}X}
\newcolumntype{L}{>{\raggedright\arraybackslash}X}

\usepackage{float}
\usepackage{url}
\usepackage{hyperref}

\usepackage{algorithm}
\usepackage{algpseudocode}
\usepackage{amssymb}
\newfloat{algorithm}{t}{lop}
\algnewcommand\algorithmicsymbols{\textbf{Symbols:}}
\algnewcommand\ASymbols{\item[\algorithmicsymbols]}
\algnewcommand\algorithmicinput{\textbf{Input:}}
\algnewcommand\algorithmicoutput{\textbf{Output:}}
\algnewcommand\Input{\item[\algorithmicinput]}
\algnewcommand\Output{\item[\algorithmicoutput]}
\algnewcommand\algorithmicforeach{\textbf{for each}}
\algdef{S}[FOR]{ForEach}[1]{\algorithmicforeach\ #1\ \algorithmicdo}
\algdef{SE}{Unless}{EndUnless}[1]{\textbf{unless} \(\mbox{#1}\) \textbf{do}}{\textbf{end unless}}
\usepackage{amsmath}

\usepackage{makecell}
    
\begin{document}

\title{Cross-Modality Clustering-based Self-Labeling \\
for Multimodal Data Classification
}

\author{\IEEEauthorblockN{1\textsuperscript{st} Paweł Zyblewski \orcidlink{0000-0002-4224-6709}}
\IEEEauthorblockA{\textit{Faculty of Information and Communication Technology}\\
\textit{Wrocław University of Science and Technology}\\
Wrocław, Poland \\
pawel.zyblewski@pwr.edu.pl}
\and
\IEEEauthorblockN{2\textsuperscript{nd} Leandro L. Minku \orcidlink{0000-0002-2639-0671}}
\IEEEauthorblockA{\textit{School of Computer Science} \\
\textit{University of Birmingham}\\
Birmingham, United Kingdom \\
l.l.minku@bham.ac.uk}
}

\maketitle

\begin{abstract}
Technological advances facilitate the ability to acquire multimodal data, posing a challenge for recognition systems while also providing an opportunity to use the heterogeneous nature of the information to increase the generalization capability of models. An often overlooked issue is the cost of the labeling process, which is typically high due to the need for a significant investment in time and money associated with human experts. Existing semi-supervised learning methods often focus on operating in the feature space created by the fusion of available modalities, neglecting the potential for cross-utilizing complementary information available in each modality. To address this problem, we propose \emph{Cross-Modality Clustering-based Self-Labeling} (\textsc{cmcsl}). Based on a small set of pre-labeled data, \textsc{cmcsl} groups instances belonging to each modality in the deep feature space and then propagates known labels within the resulting clusters. Next, information about the instances' class membership in each modality is exchanged based on the Euclidean distance to ensure more accurate labeling. Experimental evaluation conducted on 20 datasets derived from the MM-IMDb dataset indicates that cross-propagation of labels between modalities -- especially when the number of pre-labeled instances is small -- can allow for more reliable labeling and thus increase the classification performance in each modality.
\end{abstract}


\section{Introduction}
Multimodal data classification systems combine extraction methods from the fields of computer vision, natural language processing, audio processing, and tabular data analysis \cite{bayoudh2022survey}. Excellent examples are systems designed to analyze digital signals in the form of video recordings, which, unlike static images, carry an additional pool of information based on successive frames. They are the ones that often form the basis of multimodal learning, in which a visual, audio, or -- based on recording transcription -- textual layer can be extracted from a video sequence. A set of complementary, heterogeneous features extracted from multiple modalities can enable the construction of a high-quality classifier by obtaining complete information about the context of the task at hand \cite{akbari2021vatt}.

\begin{figure}[!htb]
    \centering
    \includegraphics[width=.99\columnwidth]{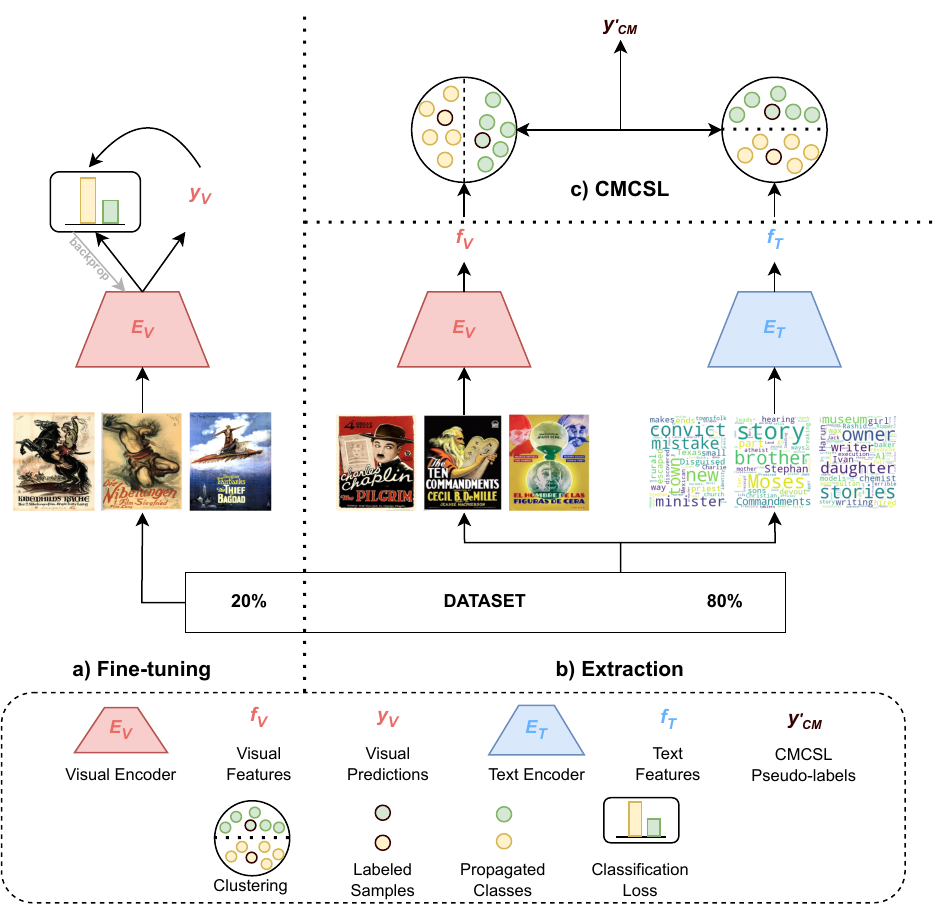}
    \caption{The general scheme of \textsc{cmcsl} along with the procedure of encoder fine-tuning and feature extraction from individual modalities.}
    \label{fig:schema}
\end{figure}

Current research and publicly available datasets in this area include sarcasm and humor detection \cite{hasan2019ur}, sentiment and subjectivity analysis \cite{zadeh2016mosi, zadeh2018multimodal}, and action recognition \cite{kay2017kinetics}. Other examples of applications include medical diagnostics \cite{johnson2016mimic} and robotics \cite{lee2019making}. Access to vast amounts of data and greater computing power encourages the use of deep learning methods. They form the basis of multimodal data analysis, which relies heavily on computer vision, where deep methods often outperform classical approaches. As confirmed by numerous scientific publications, convolutional networks are successfully used for video and audio classification (e.g., in spectrogram form) \cite{satt2017efficient} and natural language analysis \cite{gimenez2020semantic} both in the unimodal and multimodal scenario. Deep networks also enable transfer learning, allowing models to apply previously acquired knowledge to the problem at hand \cite{zhuang2020comprehensive}.

Labeling cost is a crucial issue related to real-life classification tasks \cite{gao2020consistency}. Literature on multimodal data analysis often overlooks this issue, even though label acquisition requires a lot of time and is often associated with significant financial expenses. One way to deal with this problem is a particular case of semi-supervised learning in the form of active learning \cite{settles2009active}. It allows the algorithm to automatically select the instances of interest, i.e., those potentially carrying the most helpful information about the problem under analysis, and then pass them to a domain expert for labeling. Increased focus on selecting training data instead of extending architectures and datasets is a currently developing trend in deep learning. It aims to reduce training time and labeling costs while avoiding model degradation. Other approaches to dealing with this problem include self-labeling \cite{kozal2023combining}, which focuses on methods that enable the recognition system to autonomously label instances based on a small pool of pre-labeled samples, and self-learning, in which the model is trained entirely without the labels obtained from a human expert \cite{caron2018deep}.

Even though this kind of approaches have also been applied to multimodal data, they operate mainly in the fused feature space. Additionally, the topic of semi-supervised learning for multimodal data is discussed relatively rarely, even in surveys \cite{bayoudh2022survey}. Moreover, recent literature on fully supervised multi-modal learning has shown that uni-modal networks outperform the multi-modal late fusion networks due to the modalities competing with each other \cite{huang2022modality}. These observations were made for different types of classification tasks and various combinations of modalities, even though intuitively increasing the pool of available data should allow for more information and thus a better generalization of knowledge \cite{wang2020makes}. Based on these findings, it can be assumed that in the case of limited label access and semi-supervised learning, focusing on individual modalities and then a skillful combination of the obtained information may allow for more accurate labeling and, thus, better classification quality.

This article aims to answer the question \emph{whether the exchange of information between the two modalities during the self-labeling process can lead to pseudo-labels that allow for obtaining classifiers with a greater generalization ability within individual modalities than in the case where this information exchange does not occur.} To this end, we propose \emph{Cross-Modality Clustering-based Self-Labeling} (\textsc{cmcsl}), a new method of self-labeling dedicated to multimodal data. While current \emph{state-of-the-art} methods focus on analyzing multimodal data in a feature space formed from a combination of available modalities, we employ a label cross-propagation between available modalities to improve the generalization ability of classifiers in each modality separately. To do this, \textsc{cmcsl} clusters the individual modalities based on pre-labeled instances and then propagates the known labels within the defined clusters. Then -- in case of disagreement between the labels obtained across available modalities -- the final label of a given sample is decided based on its Euclidean distance from the nearest centroid in each modality. 

In brief, the main contributions of this work are as follows:
\begin{itemize}
    \item Proposal of the \emph{Cross-Modality Clustering-based Self-Labeling} (\textsc{cmcsl}), the novel approach to self-labeling for multimodal data by employing the Euclidean distance-based label propagation between modalities.
    \item Analysis of the data preprocessing impact on the \textsc{cmcsl} performance, as it is a crucial element affecting the distance method used for label propagation.
    \item Demonstrating -- by the extensive experimental evaluation of the proposed approach on real-life MM-IMDb multimodal dataset -- the effectiveness of using the label propagation between modalities in the self-labeling process for multimodal data, thereby expanding the research field of semi-supervised learning.
\end{itemize}

\section{Related works}
In this section, we briefly describe the two most critical -- although scarcely represented in the literature -- research areas addressed in this article, i.e., multimodal data classification with limited label access and preprocessing for data clustering. It is also worth mentioning that despite the significant impact of the cost of labels on real-world data classification problems, a recent survey on multimodal learning does not include papers on semi-supervised or self-supervised classification \cite{bayoudh2022survey}. The last subsection summarizes the critical gaps in the literature and motivates our proposition.

\subsection{Classifying multimodal data with missing labels}
Very few approaches have been proposed in the literature to classify multimodal data in the case of limited label access. Among the semi-supervised learning methods, we can distinguish work by Budnik et al. \cite{budnik2014automatic}, who propose an active learning approach based on the propagation of labels in multimodal clusters using various annotation candidate selection methods. Guillaumin et al. proposed a classic self-labeling approach, in which the training set is extended by labeling samples based on the decision of \emph{Multiple Kernel Learning} (\textsc{mkl}), trained previously on the available labeled data \cite{guillaumin2010multimodal}. The \emph{Adaptive multimodal semi-supervised classification} (\textsc{ammss}) \cite{cai2013heterogeneous}, proposed by Cai et al., employs graph-based semi-supervised learning with label propagation. Zhang et al. proposed a self-labeling method using labels obtained by the classifier, which additionally considers the predictions' confidence to abstain from including uncertain samples \cite{zhang2016enhanced}.

Another popular approach to semi-supervised multimodal analysis is using an autoencoder learned on unlabeled data to optimize the network parameters and avoid overfitting \cite{kim2017multi}. Zhang et al. also employed autoencoders trained using features extracted from unlabeled samples from one or multiple joint modalities for sentiment analysis \cite{zhang2020multi}. Another example is semi-supervised learning for 3D objects using encoders and three distinct loss functions, described by Chech et al. \cite{chen2021multimodal}.

The most popular multimodal self-learning algorithm is the \emph{DeepCluster} \cite{caron2018deep} proposed by Caron et al., which uses labels obtained from $k$-means clustering for backpropagation. Based on this, Alwassel et al. proposed various approaches to cross-modal audio-video clustering \cite{alwassel2020self}. Another interesting approach, proposed by Karlos et al., is to combine self-learning and active learning \cite{karlos2019combining}.

\subsection{Preprocessing for data clustering}
Due to the fact that the method proposed in this article propagates labels based on data clustering, it is important to perform appropriate data preprocessing. The impact of the preprocessing selection on distance algorithms is a well-known and frequently discussed topic in the literature in the context of unimodal learning \cite{milligan1988study}.

Mohamad and Usman analyzed the impact of three approaches to standardization, namely Z-score, Min-Max, and decimal scaling on the $k$-means algorithm \cite{mohamad2013standardization}, showing the need to select the appropriate standardization method depending on the characteristics of the analyzed datasets. Yand et al. employed L$_2$ feature normalization for dynamic graph embedding in order to rescale all graph nodes to hypersphere \cite{yang2020featurenorm}. Aytekin et al. showed that using L$_2$ normalization constraint during auto-encoder training allows for obtaining separable and compact representations in the Euclidean space \cite{aytekin2018clustering}. The MIC (Multi-Incomplete-view Clustering) algorithm, proposed by Shao et al. \cite{shao2015multiple}, is based on weighted non-negative matrix factorization with $L_{2,1}$ regularization. Caron et al. employed $L_2$ normalization and \emph{Principal Component Analysis} (\textsc{pca}) for $k$-means clusterization in \emph{DeepCluster}. \emph{Cross-Modal Deep Clustering} proposed by Alwassel et al. also employed $L_{2}$ normalization for clustering-based self-learning \cite{alwassel2020self}.

The cited works confirm the need to analyze the impact of data preprocessing on the performance of distance algorithms depending on the characteristics of the pattern recognition task. It should be noted that the importance of this step potentially increases in the case of multimodal data, i.e., where clustering is performed in several spaces of deep features obtained as a result of independent extraction processes for each of the modalities. In this case, it is crucial to carry out preprocessing to ensure comparability of the distributions of the obtained deep feature spaces to equalize the importance of individual modalities in the cross-modal information propagation process.

\subsection{Gaps in the literature \& motivation}
The literature review showed a shortage of methods for classifying multimodal data under limited label access scenarios. In addition, the existing solutions focus primarily on analyzing multimodal data in a jointed deep feature space, resulting from a combination of feature spaces obtained in independent extraction processes for each of the available modalities. At the same time, parallel studies point to lowering the classification quality of the classifiers trained based on the combination of features extracted from individual modalities compared to its unimodal counterpart. This suggests that operating in a jointed deep feature space is not ideal. Therefore, we are conducting an extended analysis of the impact of cross-modal information transfer on classifiers trained independently within the available modalities. In this scenario, models, instead of learning on a dataset that combines all modalities, focus on one of them. Still, they use appropriately selected information from other available feature spaces in the learning process. 

To enable dealing with a limited number of labeled examples, we conduct this study in the context of semi-supervised multimodal learning via self-labeling. Our approach is called \emph{Cross-Modality Clustering-based Self-Labeling} algorithm. It allows the use of heterogeneous information contained in the available modalities to obtain pseudo-labels, which are then used to train pattern recognition models within individual modalities. The \emph{Cross-Modality Clustering-based Self-Labeling} proposal was inspired mainly by the \emph{Automatic Propagation of Manual Annotations} (\textsc{apma}) \cite{budnik2014automatic} and \emph{Cross-Modal Deep Clustering} (\textsc{xdc}) \cite{alwassel2020self}. However, at the same time, it proposes a novel approach to self-labeling of multimodal data. The differences between the literature approaches and our proposal can be summarized as follows:

\begin{itemize}
    \item \textsc{cmcsl} vs \textsc{apma}:
    \begin{itemize}
        \item \textsc{apma} proposes an active learning scenario with label increment from a human expert, while \textsc{cmcsl} offers a self-labeling approach.
        \item \textsc{apma} uses classic approaches to extraction from audio and visual modalities, while \textsc{cmcsl} employs deep neural networks.
        \item \textsc{apma} uses agglomerative clustering and then constructs multimodal clusters based on the probabilistic output of \textsc{mlp} classifers understood as the distance between modalities. \textsc{cmcsl} performs clustering independently in each modality using a single iteration of the $k$-means algorithm.
        \item \textsc{apma} only studies the case of using all pseudo-labels from a given modality to train a classifier in another one. \textsc{cmcsl} offers a dynamic cross-modal label propagation approach, allowing for a simultaneous increase in classification quality in each modality.
    \end{itemize}
    \item \textsc{cmcsl} vs \textsc{xdc}:
    \begin{itemize}
        \item \textsc{xdc} offers a clustering-based self-learning approach for deep neural networks, while \textsc{cmcsl} offers a self-labeling approach.
        \item \textsc{xdc} is applicable only in the initial training of neural networks on large datasets for transfer learning purposes.
        \item \textsc{xdc}, despite the cross-use of labels obtained in different modalities, at no point proposes a dynamic approach to label selection, taking into account the possibility of different labeling of individual instances in different modalities.
    \end{itemize}
\end{itemize}

To our knowledge, \textsc{cmcsl} is the first attempt to apply cross-modal label propagation in the task of self-labeling multimodal data while also proposing a mechanism of dynamic selection of pseudo-labels from the pool available in different modalities. Additionally, this work verifies the preliminary conclusions obtained by Budnik et al. \cite{budnik2014automatic}. It also significantly expands their research by taking on a different pattern recognition task and verifying the proposed novel approach using an extended and robust experimental protocol.

\section{Cross-Modality Clustering-based Self-Labeling}
The general scheme of applying the \textsc{cmcsl} for the MM-IMDb multimodal dataset used in this work is presented in Fig. \ref{fig:schema}, including the extraction process of deep features from available visual and text modalities. The \emph{Cross-Modality Clustering-based Self-Labeling} is based on the assumption that individual modalities carry heterogeneous, complementary information. Thanks to this, the distributions of instances belonging to particular problem classes in the feature space may be more concentrated, allowing better separability of classes in some areas of the feature space \cite{budnik2014automatic}. This could potentially enable more accurate clustering and, thus, more reliable pseudo-labeling based on the propagation of the known labels.

First, deep features are extracted from a dataset using appropriate architectures. In this case, ResNet-18 architecture (pretrained on ImageNet) is used for visual and MiniLM-L6-v2 for text modality. Following the procedure shown in Fig. \ref{fig:schema}, each dataset is split in the ratio 20:80, where 20\% is used for ResNet-18 fine-tuning. These proportions have been selected to enable fine-tuning of the visual encoder while retaining as much data as possible to evaluate the proposed approach. The weights corresponding to the ratio of class cardinality are included in the fine-tuning process to offset the data imbalance. From the remaining 80\% of the data, deep features are extracted separately for both modalities. After extraction, the visual and text modalities contained 512 and 384 features, respectively. Then, the main steps of the \textsc{cmcsl} algorithm, presented in Pseudocode \ref{alg:cmcsl}, are performed. In the pseudocode we can disntinguish the following methods:
\begin{itemize}
    \item $\textsc{Labeling}()$ -- labels $b_{class}$ instances, randomly selected based on uniform distribution, from each of the $n_{classes}$ classes of multimodal dataset $\mathcal{X}$. These instances are treated as pre-labeled centroids $\mathcal{M}$ for clustering purposes.
    \item $\textsc{Preprocess}()$ -- applies the chosen data preprocessing method to the given modality $X$,
    \item $\textsc{Clustering}()$ -- groups a given single modality $X$ into a set of clusters $\mathcal{C}$, containing $c=n_{classes} * b_{class}$ clusters. Pre-labeled samples $\mathcal{M}$ are used as initial centroids for a single iteration of the $k$-means algorithm. Clusters are defined once and no new centroids are determined.
    \item $\textsc{Propagate}()$ -- propagates classes of the pre-labeled centroids $\mathcal{M}$ to all instances belonging to their corresponding clusters $C$.
    \item $\textsc{Nearest}()$ -- if the labels of a given $x_i$ instance in both modalities differ, the Euclidean distance of the $x_i$ instance from the centroid of its cluster in each modality is checked. The label from the modality where $x_i$ is closer to the centroid is considered correct.
\end{itemize}

\begin{algorithm}[!htb]
\small
\caption{Pseudocode of the proposed \textsc{cmcsl} algorithm.}
\label{alg:cmcsl}
\begin{algorithmic}[1]
\Input
\Statex $\mathcal{X}=\{X_V, X_T\}$ -- set of modalities.
\Statex $n_{classes}$ -- number of problem classes,
\Statex $b_{class}$ -- number of pre-labeled samples in each class,
\ASymbols
\Statex $X_V=\{x^{V}_1, x^{V}_2, \ldots, x^{V}_N\}$ -- visual modality,
\Statex $X_T=\{x^{T}_1, x^{T}_2, \ldots, x^{T}_N\}$ -- text modality,
\Statex $N$ -- number of samples,
\Statex $d_V$ -- visual modality dimensionality,
\Statex $d_T$ -- text modality dimensionality,
\Statex $c=n_{classes} * b_{class}$ -- number of clusters,
\Statex $\mu$ -- centroid,
\Statex $\mathcal{M}=\{(\mu_1, y_1), (\mu_2, y_2), \ldots, (\mu_c, y_c)\}$ -- labeled centroids,
\Statex $\mathcal{C}_V=\{C^V_1, C^V_2, \ldots, C^V_c\}$ -- set of visual modality clusters.
\Statex $\mathcal{C}_T=\{C^T_1, C^T_2, \ldots, C^T_c\}$ -- set of text modality clusters,
\Statex $y'_V, y'_T$ -- visual and text pseudo-labels.
\Output d
\Statex $y'_{CM}$ -- cross-modal pseudo-labels.
    \vspace{1em}
    \State $y'_{CM} \gets \varnothing$
    \State $\mathcal{M} = \textsc{Labeling}(\mathcal{X}, n_{classes}, b_{class})$
    \State $X_V, X_T = \textsc{Preprocess}(X_V), \textsc{Preprocess}(X_T)$
    \State $\mathcal{C}_V,\mathcal{C}_T=\textsc{Clustering}(X_V, \mathcal{M}),\textsc{Clustering}(X_T, \mathcal{M})$
    \State $y'_V, y'_T = \textsc{Propagate}(C_V, \mathcal{M}),  \textsc{Propagate}(C_T, \mathcal{M})$
    \ForEach{$label^V_i, label^T_i$ in $y'_V, y'_T$}
        \If{$label^V_i == label^T_i$}
            \State $y'_{CM} \gets label^V_i$ 
        \Else
            \State $y'_{CM} \gets \textsc{Nearest}(x^{V}_i, label^V_i, x^{T}_i, label^T_i, \mathcal{M})$
        \EndIf
    \EndFor
\end{algorithmic}
\end{algorithm}

As input, \textsc{cmcsl} receives data in the form of available modalities. In the case of this work, it is the visual modality $X_V$ and the text modality $X_T$, as well as the number of problem classes $n_{classes }$ and the labeling budget $b_{class}$ for each class, which tells us how many instances belonging to each class should be pre-labeled. The step-by-step description is as follows. First, \textsc{cmcsl} obtains the labels for randomly selected $b_{class}$ instances from each of the $n_{classes}$ classes (Line 2, $\textsc{Labeling}()$ method). These instances will serve as labeled centroids of the $\mathcal{M}$ clusters of $\mathcal{C_V}$ and $\mathcal{C_T}$ that will be created independently in both modalities. Before the clustering process, each modality is preprocessed to bring the feature space of individual modalities to a comparable representation (Line 3, $\textsc{Preprocess}()$ method). Then, in each of the modalities, the clustering process is carried out using the previously labeled examples as initial centroids (Step 4 $\textsc{Clustering}()$ method). This is adequate to a single iteration of the $k$-means algorithm -- often used in deep features clusterization \cite{caron2018deep, alwassel2020self} -- with predefined centroids. Then, the centroid classes are propagated to all instances belonging to their respective clusters (Line 5, $\textsc{Propagate}()$ method). The next step is to compare the labels that $x_i$ received in both $X_V$ and $X_T$ modalities (Line 6). If label is the same in both modalities, we consider it correct (Line 7 \& 8). In the case where the labels differ, the one from the modality where $x_i$ is closer to its centroid in terms of the Euclidean distance is assumed to be correct (Line 9 \& 10, $\textsc{Nearest}()$ method) and used for both modalities. After this process, a classifier is trained for each modality based on the data in the given modality and pseudo-labels obtained as a result of cross-propagation (same for each modality).

\section{Experimental evaluation}
In this section, we describe in detail the experimental study conducted to investigate the properties of \textsc{cmcsl}. The experiments were designed to answer the following research questions:
\begin{itemize}
    \item \textbf{RQ1} What is the impact of data preprocessing on \textsc{cmcsl} performance?
    \item \textbf{RQ2} Can the exchange of information between modalities in the self-labeling process enable a classifier with generalization ability superior to a model learned solely from labels obtained within a single modality?
    \item \textbf{RQ3} How much influence the used classification algorithm has on the benefit resulting from the cross-propagation of labels between modalities?
\end{itemize}

\subsection{Set-up}

\noindent\textbf{Data}
\begin{figure*}[!htb]
    \centering
    \includegraphics[width=.9\textwidth]{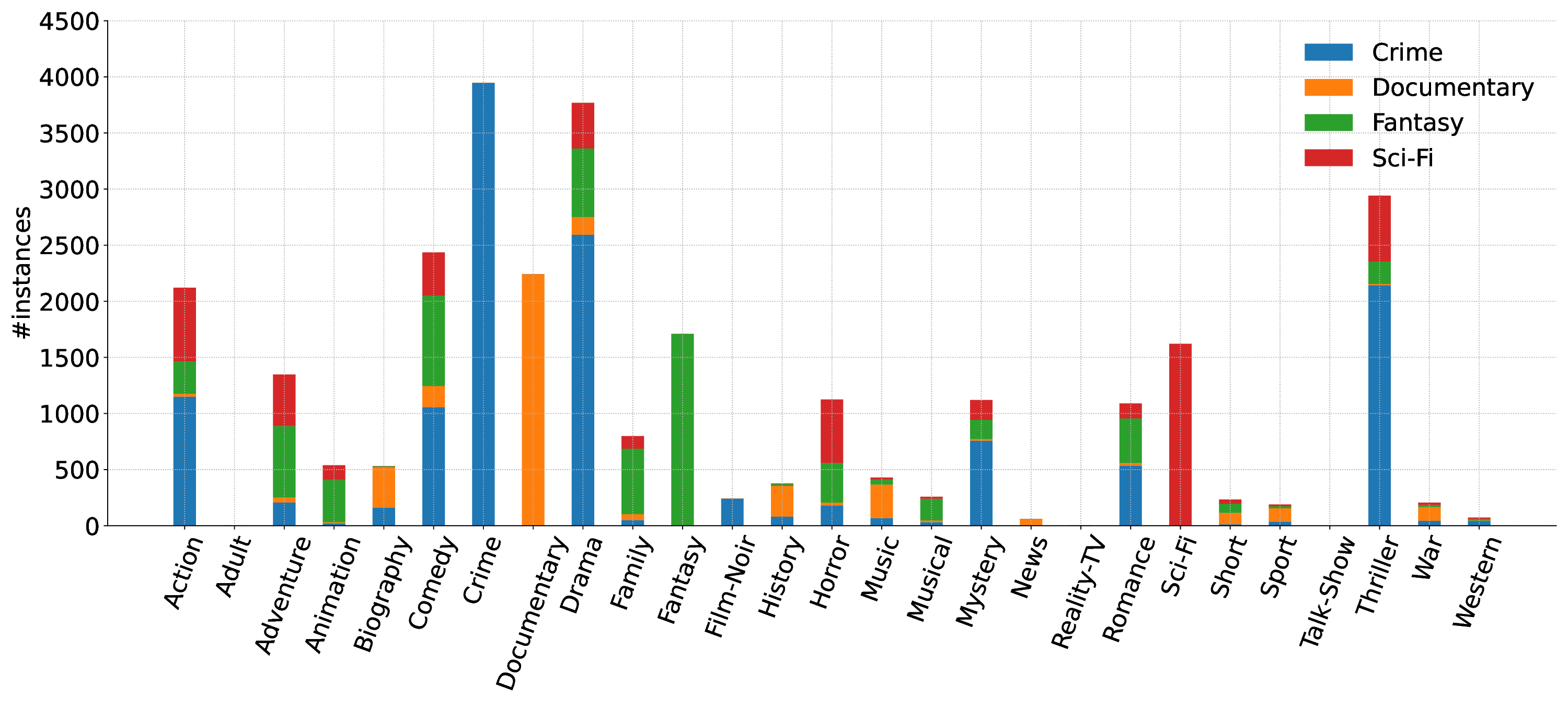}
    \caption{Visualization of the label distribution for a four-class \textsc{cdfs} (Crime, Documentary, Fantasy and Sci-Fi) subset after the transition from multilabel to multiclass. The final subset is composed only of selected, non-overlapping classes.}
    \label{fig:cdfs}
\end{figure*}
Experiments were conducted on datasets defined as subsets of the MM-IMDb dataset \cite{arevalo2017gated}, which presents a multilabel movie genre classification problem with visual (movie posters) and text (plot summaries) modalities. For the purposes of research, 20 subsets named with acronyms composed of the first letters of the contained classes were separated from MM-IMDb, 10 of which contain binary and 10 multiclass problems. The division was made so that the labels of a given subset were unique, i.e., they did not overlap. Fig. \ref{fig:cdfs} shows the distribution of classes of the \textsc{cdfs} subset (Crime, Documentary, Fantasy, Sci-Fi) obtained in this way, where each of the selected classes is unique in relation to the others, and the remaining labels available in the subset are their composition. Table \ref{tab:data} presents the characteristics of all obtained subsets. While the MM-IMDb dataset can be interpreted as a data stream \cite{niedziolka2023non}, this work focuses on introducing \textsc{cmcsl} and testing it for static data, possibly applying it to data stream in future work.

\begin{table}[!htb]
    \centering
    \scriptsize
    \caption{Characteristics of the data subsets extracted from the MM-IMDb dataset.}
    \renewcommand{\arraystretch}{0.7}
\begin{tabularx}{0.99\columnwidth}{llc}
\toprule
 Abbreviation   & Classes (class samples)                                                                                                                    &   
 \makecell{Total \\samples} \\
\midrule
 \textbf{HR}             & \textbf{H}orror (2151), \textbf{R}omance (4653)                                                                                                       &       6804 \\
 \midrule
 \textbf{DS}             & \textbf{D}ocumentary (1870), \textbf{S}ci-Fi (1697)                                                                                                   &       3567 \\
 \midrule
 \textbf{HM}             & \textbf{H}istory (1005), \textbf{M}usic (883)                                                                                                         &       1888 \\
 \midrule
 \textbf{BW}             & \textbf{B}iography (1164), \textbf{W}estern (593)                                                                                                     &       1757 \\
 \midrule
 \textbf{AM}             & \textbf{A}nimation (682), \textbf{M}usical (560)                                                                                                      &       1242 \\
 \midrule
 \textbf{FS}             & \textbf{F}ilm-Noir (266), \textbf{S}hort (474)                                                                                                        &        740 \\
 \midrule
 \textbf{FW}             & \textbf{F}amily (1479), \textbf{W}ar (1139)                                                                                                           &       2618 \\
 \midrule
 \textbf{MS}             & \textbf{M}usical (709), \textbf{S}port (543)                                                                                                          &       1252 \\
 \midrule
 \textbf{FM}             & \textbf{F}antasy (1556), \textbf{M}ystery (1638)                                                                                                      &       3194 \\
 \midrule
 \textbf{AT}             & \textbf{A}dventure (1983), \textbf{T}hriller (3924)                                                                                                   &       5907 \\
 \midrule
 \textbf{HMSWW}          & \makecell[l]{\textbf{H}istory (682), \textbf{M}usical (657), \textbf{S}ci-Fi (1642), \\\textbf{W}ar (770), \textbf{W}estern (524)}                                                               &       4275 \\
 \midrule
 \textbf{FMSW}           & \makecell[l]{\textbf{F}amily (1059), \textbf{M}usical (448), \textbf{S}ci-Fi (1486), \\\textbf{W}ar (1105)}                                                                             &       4098 \\
 \midrule
 \textbf{FMNSSW}         & \makecell[l]{\textbf{F}ilm-Noir (260), \textbf{M}usical (664), \textbf{N}ews (54), \\\textbf{S}hort (442), \textbf{S}port (533), \textbf{W}estern (587)}                                                  &       2540 \\
 \midrule
 \textbf{FFW}            & \textbf{F}antasy (1729), \textbf{F}ilm-Noir (265), \textbf{W}estern (602)                                                                                      &       2596 \\
 \midrule
 \textbf{CDFS}           & \makecell[l]{\textbf{C}rime (3157), \textbf{D}ocumentary (1793), \textbf{F}antasy (1368), \\\textbf{S}ci-Fi (1297)}                                                                     &       7615 \\
 \midrule
 \textbf{AFHMS}          & \makecell[l]{\textbf{A}nimation (394), \textbf{F}antasy (1151), \textbf{H}istory (974), \\\textbf{M}ystery (1577), \textbf{S}port (505)}                                                         &       4601 \\
 \midrule
 \textbf{ACMD}           & \makecell[l]{\textbf{A}dventure (2190), \textbf{C}rime (3053), \textbf{M}usic (598), \\\textbf{D}ocumentary (1535)}                                                                     &       7376 \\
 \midrule
 \textbf{ACCDDHMRS}      & \makecell[l]{\textbf{A}ction (371), \textbf{C}omedy (1885), \textbf{C}rime (91), \\\textbf{D}ocumentary (1496), \textbf{D}rama (3802), \textbf{H}orror (583), \\\textbf{M}ystery (107), \textbf{R}omance (127), \textbf{S}ci-Fi (118)} &       8580 \\
 \midrule
 \textbf{ABHMW}          & \makecell[l]{\textbf{A}nimation (778), \textbf{B}iography (653), \textbf{H}istory (437), \\\textbf{M}usic (694), \textbf{W}ar (749)}                                                             &       3311 \\
 \midrule
 \textbf{ABFHSS}         & \makecell[l]{\textbf{A}ction (2509), \textbf{B}iography (987), \textbf{F}amily (1097), \\\textbf{H}orror (1896), \textbf{S}hort (295), \textbf{S}port (332)}                                              &       7116 \\
\bottomrule
\end{tabularx}
    \label{tab:data}
\end{table}

\noindent\textbf{Experimental protocol.}
In order to ensure a robust experimental protocol, all experiments were conducted according to the 5 times repeated 2-fold stratified cross-validation, based only on the data used for deep feature extraction ($80\%$ of each dataset). The obtained results were supported by the Combined 5x2 CV F-test \cite{alpaydm1999combined} and Wilcoxon signed-rank test \cite{stapor2021design} with $\alpha=0.05$ (the higher the rank, the better). The Combined 5x2 CV F-test is used to analyze the statistical relationships between methods within individual datasets, while the Wilcoxon signed-rank test enables a global comparison of the tested approaches (separately within binary and multiclass datasets). The classification performance was evaluated based on the \emph{balanced accuracy score} (\textsc{bac}).

\noindent\textbf{Reproducibility.}
All experiments presented in this paper were carried out in \emph{Python} using the \emph{scikit-learn} library \cite{scikit-learn}, while the extraction of deep features from datasets was performed using the \emph{PyTorch} library \cite{paszke2017automatic}. The self-labeling methods used were tested for three base classifiers, namely \emph{Gaussian Na\"ive Bayes} (\textsc{gnb}), \emph{Logistic Regression} (\textsc{lr}), and \emph{Classification and Regression Tree} (\textsc{cart}), in \emph{scikit-learn} library implementations. Two \emph{GitHub} repositories have been made available to enable the replication of all the results reported in this paper. The first one\footnote{\url{https://github.com/w4k2/mm-datasets}} contains the code responsible for dividing the MM-IMDb dataset into subsets and their extraction into a tabular form (downloading the publicly available MM-IMDb set is required). The second repository\footnote{\url{https://github.com/w4k2/cmcsl}} contains the implementation of all the conducted experiments regarding \textsc{cmcsl}. The second repository also contains all the figures and the complete versions of the tables, including the results that were omitted from the article.

\subsection{Experiment scenarios}

\noindent\textbf{Experiment 1 -- Classification algorithm selection}
The first experiment aims to investigate which of the three base classifiers achieves the statistically significantly best performance in terms of \emph{balanced accuracy score} for individual modalities of each of the 20 datasets under full labeling. Due to the deep feature extraction process carried out previously for each dataset, this experiment is equivalent to determining which classification algorithm should replace the decision layer of the deep neural network. In the next two experiments, full results will be presented for the selected classifier, while the results achieved by the other two models will be discussed in abbreviated form.

\noindent\textbf{Experiment 2 -- Preprocessing impact}
The second experiment aims to analyze the impact of data preprocessing approaches widely used in clustering methods on the performance of a classifier trained based on \textsc{cmcsl} pseudo-labels. Due to the fact that the proposed algorithm analyzes the Euclidean distance in two separate feature spaces, it is necessary to reduce them to a comparable representation. The experiment was carried out for the value of the $b_{class}$ parameter in the range of 1-20. To this end, the following preprocessing approaches were explored: 
\begin{itemize}
    \item \emph{No data preprocessing} (\textsc{raw}) -- both modalities preserved in the form obtained in the extraction process.
    \item \emph{L$_2$ Normalization} ($L2$) -- an application of L$_2$ normalization to both modalities so that the sum of the squares of each instance is 1.
    \item \emph{Standard Scaling} (\textsc{std}) -- standardizing features of both modalities so that mean of observed values is 0 and the standard deviation is 1.
    \item \emph{MinMax Scaling} (\textsc{mm}) -- scaling the features of both modalities to $0-1$ range.
    \item Combination of L2 and \textsc{std} (\textsc{l2std}).
\end{itemize}

\noindent\textbf{Experiment 3 -- Comparison of \textsc{cmcsl} with reference methods}
The aim of the third experiment is to compare the classification performance of a classifier built using pseudo-labels acquired using \textsc{cmcsl} algorithm with following reference approaches: 
\begin{itemize}
    \item \emph{Full labeling} (\textsc{full}) -- classification algorithm trained on fully labeled data.
    \item \emph{Early Fusion} (\textsc{ef}) -- one classifier trained on concatenated spaces, preprocessed using the approach selected in Experiment 2.
    \item \emph{Late Fusion} (\textsc{lf}) -- support accumulation based combination of classifiers trained on both modalities assuming full labeling.
    \item \emph{Pre-labeled only} (\textsc{pre}) -- classification algorithm trained only on the pre-labeled samples.
    \item \emph{Unimodal clustering-based label propagation} (\textsc{uni}) -- classification algorithm trained on the pseudo-labels obtained by the standard propagation of labels in each cluster of a given modality without cross-modal information exchange.
\end{itemize}
Again, the experiment was carried out for the value of the $b_{class}$ parameter in the range of 1-20. It is worth emphasizing that no comparison was made with any other self-labeling algorithm based on cross-modal label propagation, as the literature review did not show the existence of approaches comparable to the proposed \textsc{cmcsl}.

\subsection{Experiment 1 -- Classification algorithm selection}
Table \ref{tab:ex1_short} shows the Wilcoxon global rank test results for the three base classifiers for binary and multiclass datasets under full labeling. 
The first row shows the averaged classifier ranks in terms of \emph{balanced accuracy score}. The indices (by column) of the algorithms from which the given model is statistically significantly better are given below.

Based on the obtained results, it can be concluded that the extraction potential of the applied deep network architectures led to independent and uniform deep features. This resulted in \textsc{gnb} allowing for the most coherent separation of categories, obtaining statistically significantly the best result for multiclass data while being comparable to \textsc{lr} in the case of binary problems. Therefore, the \textsc{gnb} classifier was chosen for subsequent experiments.
\begin{table}[!htb]
    \centering
    \scriptsize
    \caption{Results of Wilcoxon statistical test ($\alpha=0.05$) for base classifiers trained on fully labeled data. 
    The first row shows the averaged classifier ranks in terms of \textsc{bac}. The indices (by column) of the algorithms from which the given model is statistically significantly better are given below the rank.
    }
    \renewcommand{\arraystretch}{0.7}
\begin{tabularx}{0.99\columnwidth}{ll|CCC}
\toprule
      &    & GNB$^1$   & LR$^2$   & CART$^3$   \\
\midrule
\multicolumn{2}{l}{Binary} & 2.400 & \textbf{2.600}  & 1.000  \\
              &            & 3     & 3      & ---    \\
\midrule
\multicolumn{2}{l}{Multiclass} & \textbf{2.750} & 2.250  & 1.000  \\
              &            & all   & 3      & ---    \\
\bottomrule
\end{tabularx}
    \label{tab:ex1_short}
\end{table}

\subsection{Experiment 2 -- Preprocessing impact}
Tables \ref{tab:ex2_binary} and \ref{tab:ex2_multi} show the results of comparing the average balanced accuracy obtained by \textsc{gnb} trained on pseudo-labels from \textsc{cmcsl} for both modalities of binary and multiclass datasets. The presented results were averaged in terms of the $b_{class}$ parameter. We can observe that in both cases, the use of L$_2$ Normalization in combination with Standard Scaling allows us to obtain the most comparable feature spaces in both modalities, which translates into the highest rank.

\begin{figure*}[!htb]
    \centering
    \includegraphics[width=.99\textwidth]{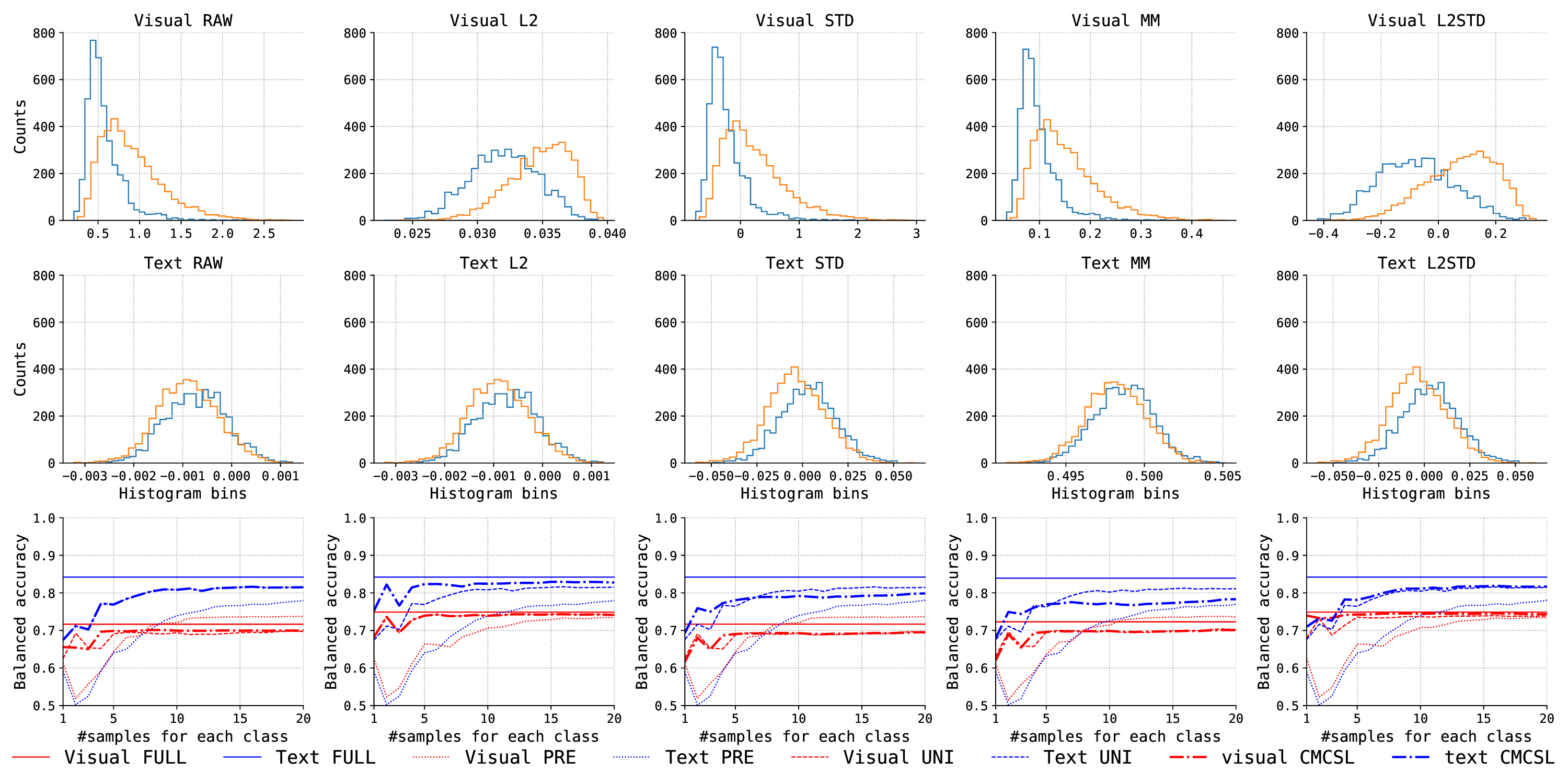}
    \caption{Visualization of the impact of data preprocessing on the distribution of classes in the space of average samples and the \emph{balanced accuracy score} of the classifer trained on \textsc{cmcsl} pseudo-labels for AT dataset. 
    }
    \label{fig:ex2_at}
\end{figure*}

\begin{table}[!htb]
    \centering
    \scriptsize
    \caption{Impact of preprocessing for binary datasets. The first row for each dataset shows the averaged \textsc{bac}. The indices (by column) of the algorithms from which the given model is statistically significantly better based on the 5x2 CV F-test ($\alpha=0.05$) are given below. The last two rows present the Wilcoxon test.
    }
    \renewcommand{\arraystretch}{0.7}
\begin{tabularx}{0.99\columnwidth}{ll|CCCCC}
\toprule
 Dataset      &M   & RAW$^1$   & L2$^2$   & STD$^3$   & MM$^4$   & L2STD$^5$   \\
\midrule
 HR           & Visual & 0.671   & \textbf{0.775}   & 0.661 & 0.673 & 0.774   \\                                                                                            
              &        & ---     & 1, 3, 4 & ---   & ---   & 1, 3, 4 \\                                                                                            
 HR           & Text   & 0.800   & \textbf{0.862}   & 0.799 & 0.802 & 0.818   \\                                                                                            
              &        & ---     & all     & ---   & ---   & 1       \\                                                                                            
 DS           & Visual & 0.758   & 0.752   & 0.744 & 0.745 & \textbf{0.771}   \\                                                                                            
              &        & 3, 4    & ---     & ---   & ---   & 1, 3, 4 \\                                                                                            
 DS           & Text   & 0.904   & 0.890   & 0.902 & 0.901 & \textbf{0.909}   \\                                                                                            
              &        & ---     & ---     & ---   & ---   & 1       \\                                                                                            
 HM           & Visual & 0.651   & 0.596   & 0.644 & 0.633 & \textbf{0.653}   \\                                                                                            
              &        & 2, 3, 4 & ---     & 4     & ---   & 2, 4    \\                                                                                            
 HM           & Text   & \textbf{0.887}   & 0.767   & 0.884 & 0.882 & 0.884   \\                                                                                            
              &        & 2       & ---     & 2     & ---   & ---     \\                                                                                            
 BW           & Visual & 0.703   & 0.740   & 0.685 & 0.699 & \textbf{0.752}   \\                                                                                            
              &        & 3       & 1, 3, 4 & ---   & 3     & 1, 3, 4 \\                                                                                            
 BW           & Text   & 0.879   & \textbf{0.889}   & 0.853 & 0.864 & 0.876   \\                                                                                            
              &        & 3, 4    & 3       & ---   & 3     & 3, 4    \\                                                                                            
 AM           & Visual & 0.813   & 0.824   & 0.807 & 0.812 & \textbf{0.832}   \\                                                                                            
              &        & 3       & 1, 3    & ---   & ---   & all     \\                                                                                            
 AM           & Text   & 0.807   & \textbf{0.830}   & 0.818 & 0.811 & 0.815   \\                                                                                            
              &        & ---     & ---     & 1, 4  & ---   & 1       \\                                                                                            
 FS           & Visual & 0.842   & \textbf{0.873}   & 0.844 & 0.864 & 0.866   \\                                                                                            
              &        & ---     & 1, 3    & ---   & 3     & 1, 3    \\                                                                                            
 FS           & Text   & 0.863   & \textbf{0.876 }  & 0.862 & 0.874 & 0.864   \\                                                                                            
              &        & ---     & ---     & ---   & 1     & ---     \\                                                                                            
 FW           & Visual & 0.742   & 0.744   & 0.736 & 0.740 & \textbf{0.761}   \\                                                                                            
              &        & 3       & ---     & ---   & ---   & 1, 3, 4 \\                                                                                            
 FW           & Text   & \textbf{0.931}   & 0.916   & 0.922 & 0.919 & 0.930   \\                                                                                            
              &        & 3, 5    & ---     & ---   & ---   & 3       \\                                                                                            
 MS           & Visual & 0.753   & 0.712   & 0.749 & 0.750 & \textbf{0.758}   \\                                                                                            
              &        & ---     & ---     & ---   & ---   & 1, 3    \\                                                                                            
 MS           & Text   & \textbf{0.913}   & 0.852   & \textbf{0.913} & 0.910 & 0.912   \\                                                                                            
              &        & ---     & ---     & ---   & ---   & ---     \\                                                                                            
 FM           & Visual & 0.675   & 0.636   & 0.636 & 0.636 & \textbf{0.676}   \\                                                                                            
              &        & 3, 4    & ---     & ---   & 3     & 3, 4    \\                                                                                            
 FM           & Text   & 0.782   & 0.744   & 0.774 & 0.772 & \textbf{0.779}   \\                                                                                            
              &        & ---     & ---     & ---   & ---   & ---     \\                                                                                            
 AT           & Visual & 0.693   & 0.735   & 0.686 & 0.691 & \textbf{0.743}   \\                                                                                            
              &        & ---     & 1, 3, 4 & ---   & ---   & 1, 3, 4 \\                                                                                            
 AT           & Text   & 0.788   & \textbf{0.818}   & 0.781 & 0.766 & 0.796   \\                                                                                            
              &        & 4       & 3, 4    & 4     & ---   & 3, 4    \\
\midrule
\multicolumn{2}{l}{Average rank} & 3.300 & 3.100           & 2.100         & 2.300    & \textbf{4.200}                         \\
              &        & 3, 4    & ---     & ---   & ---   & all     \\
\bottomrule
\end{tabularx}
    \label{tab:ex2_binary}
\end{table}

\begin{table}[!htb]
    \centering
    \scriptsize
    \caption{Impact of preprocessing for multiclass datasets.
    }
    \renewcommand{\arraystretch}{0.7}
\begin{tabularx}{0.99\columnwidth}{ll|CCCCC}
\toprule
      &   & RAW$^1$   & L2$^2$   & STD$^3$   & MM$^4$   & L2STD$^5$   \\
\midrule
\multicolumn{2}{l}{Average rank} & 4.350   & 1.250           & 2.450         & 2.450    & \textbf{4.500}                         \\
              &        & 2, 3, 4 & ---     & 2     & 2     & all     \\
\bottomrule
\end{tabularx}
    \label{tab:ex2_multi}
\end{table}
In addition, in Fig. \ref{fig:ex2_at}, we can observe how class distributions and classification performance of reference methods change in individual modalities depending on the approach to preprocessing used. The first two rows present the histograms of the averaged sample value in each class. The third row shows the \textsc{bac} achieved by \textsc{cmcsl} and reference methods depending on the value of the $b_{class}$ parameter. Based on this information, it is possible to answer \textbf{RQ1}.

The lack of preprocessing equates to the selection of labels exclusively from the text modality. This is due to the fact that the deep features obtained from the MiniLM model are already subjected to L$_2$ normalization. Therefore, the space is more compact. Performing L$_2$ normalization is equivalent to using labels exclusively from the visual modality. This is due to the fact that this modality has a higher dimensionality, so it's denser after the normalization. Interestingly, in some cases, the use of labels derived from propagation in the \emph{weaker} modality results in a significant increase in classification performance in the modality with the stronger discriminatory capacity, which can be seen in Fig. \ref{fig:ex2_at}. A \emph{weaker} modality is considered to be the one that offers a lower \emph{balance accuracy score} when the classifier is trained on fully labeled data within said modality.

The Standard Scaling alone leads to a deterioration of the results since the feature spaces are not comparable with each other. The use of Standard Scaling preceded by L$_2$ normalization allows to achieve a performance improvement in both modalities. This is due to the fact that only after carrying out both operations do the feature spaces of individual modalities begin to be comparable. 

In light of the above conclusions, L$_2$ normalization and Standard Scaling are used in the third stage. This is due to the observation that this is the only combination that allows simultaneous improvement of classifier performance in both problem modalities, especially in the case of a very low number of pre-labeled samples.

\subsection{Experiment 3 -- Comparison of \textsc{cmcsl} with reference methods}
\begin{table}[!htb]
    \centering
    \scriptsize
    \caption{Comparison with reference methods for binary datasets.
    }
   \renewcommand{\arraystretch}{0.7}
\begin{tabularx}{0.99\columnwidth}{llCCC|CCC}
\toprule
Dataset      & M   & FULL   & EF    & LF    & PRE$^1$   & UNI$^2$   & CMCSL$^3$   \\
\midrule                                                                                                                                                                                                                                  
 HR           & Visual & 0.791  & 0.802 & 0.846 & 0.720 & \textbf{0.776} & 0.774   \\                                                                                                                                                              
              &        &        &       &       & ---   & 1     & 1       \\                                                                                                                                                              
 HR           & Text   & 0.890  & 0.802 & 0.846 & 0.730 & 0.798 & \textbf{0.818}   \\                                                                                                                                                              
              &        &        &       &       & ---   & 1     & all     \\                                                                                                                                                              
 DS           & Visual & 0.773  & 0.841 & 0.868 & 0.707 & 0.747 & \textbf{0.771}   \\                                                                                                                                                              
              &        &        &       &       & ---   & ---   & 1       \\                                                                                                                                                              
 DS           & Text   & 0.930  & 0.841 & 0.868 & 0.792 & 0.904 & \textbf{0.909}   \\                                                                                                                                                              
              &        &        &       &       & ---   & 1     & 1       \\                                                                                                                                                              
 HM           & Visual & 0.665  & 0.881 & 0.881 & 0.565 & 0.598 & \textbf{0.653}   \\                                                                                                                                                              
              &        &        &       &       & ---   & ---   & all     \\                                                                                                                                                              
 HM           & Text   & 0.924  & 0.881 & 0.881 & 0.787 & \textbf{0.884} & \textbf{0.884}   \\                                                                                                                                                              
              &        &        &       &       & ---   & 1     & 1       \\                                                                                                                                                              
 BW           & Visual & 0.767  & 0.855 & 0.886 & 0.699 & 0.737 & \textbf{0.752}   \\                                                                                                                                                              
              &        &        &       &       & ---   & 1     & 1       \\                                                                                                                                                              
 BW           & Text   & 0.929  & 0.855 & 0.886 & 0.815 & 0.875 & \textbf{0.876}   \\                                                                                                                                                              
              &        &        &       &       & ---   & 1     & 1       \\                                                                                                                                                              
 AM           & Visual & 0.840  & 0.865 & 0.873 & 0.772 & 0.823 & \textbf{0.832}   \\                                                                                                                                                              
              &        &        &       &       & ---   & 1     & 1       \\                                                                                                                                                              
 AM           & Text   & 0.854  & 0.865 & 0.873 & 0.716 & 0.809 & \textbf{0.815}   \\                                                                                                                                                              
              &        &        &       &       & ---   & 1     & all     \\                                                                                                                                                              
 FS           & Visual & 0.923  & 0.953 & 0.947 & 0.831 & 0.865 & \textbf{0.866}   \\                                                                                                                                                              
              &        &        &       &       & ---   & 1     & 1       \\                                                                                                                                                              
 FS           & Text   & 0.926  & 0.953 & 0.947 & 0.801 & \textbf{0.864} & \textbf{0.864 }  \\                                                                                                                                                              
              &        &        &       &       & ---   & 1     & 1       \\                                                                                                                                                              
 FW           & Visual & 0.759  & 0.851 & 0.889 & 0.693 & 0.744 & \textbf{0.761}   \\                                                                                                                                                              
              &        &        &       &       & ---   & 1     & all     \\                                                                                                                                                              
 FW           & Text   & 0.945  & 0.851 & 0.889 & 0.822 & 0.929 & \textbf{0.930}   \\                                                                                                                                                              
              &        &        &       &       & ---   & 1     & 1       \\                                                                                                                                                              
 MS           & Visual & 0.756  & 0.852 & 0.872 & 0.672 & 0.710 & \textbf{0.758}   \\                                                                                                                                                              
              &        &        &       &       & ---   & 1     & 1       \\                                                                                                                                                              
 MS           & Text   & 0.946  & 0.852 & 0.872 & 0.791 & 0.911 & \textbf{0.912}   \\                                                                                                                                                              
              &        &        &       &       & ---   & 1     & all     \\                                                                                                                                                              
 FM           & Visual & 0.691  & 0.779 & 0.782 & 0.607 & 0.639 & \textbf{0.676}   \\                                                                                                                                                              
              &        &        &       &       & ---   & ---   & 1       \\                                                                                                                                                              
 FM           & Text   & 0.834  & 0.779 & 0.782 & 0.687 & 0.778 & \textbf{0.779}   \\                                                                                                                                                              
              &        &        &       &       & ---   & 1     & 1       \\                                                                                                                                                              
 AT           & Visual & 0.749  & 0.769 & 0.802 & 0.682 & 0.730 & \textbf{0.743}   \\                                                                                                                                                              
              &        &        &       &       & ---   & 1     & 1       \\                                                                                                                                                              
 AT           & Text   & 0.842  & 0.769 & 0.802 & 0.701 & 0.787 & \textbf{0.796}   \\                                                                                                                                                              
              &        &        &       &       & ---   & 1     & 1       \\

\midrule
\multicolumn{4}{l}{Average rank} & & 1.000 & 2.050 & \textbf{2.950}   \\
              &     &        &       &       & ---   & 1     & all     \\
\bottomrule
\end{tabularx}
    \label{tab:ex3_binary}
\end{table}

\begin{table}[!htb]
    \centering
    \scriptsize
    \caption{Comparison with reference methods for multiclass datasets.
    }
    \renewcommand{\arraystretch}{0.7}
\begin{tabularx}{0.99\columnwidth}{l|CCC}
\toprule
    & PRE$^1$   & UNI$^2$   & CMCSL$^3$   \\
\midrule
Average rank & 1.000  & 2.150    & \textbf{2.850}   \\
             & ---    & 1        & all     \\
\bottomrule
\end{tabularx}
    \label{tab:ex3_multi}
\end{table}

\begin{figure}[!htb]
    \centering
    \includegraphics[width=.99\columnwidth]{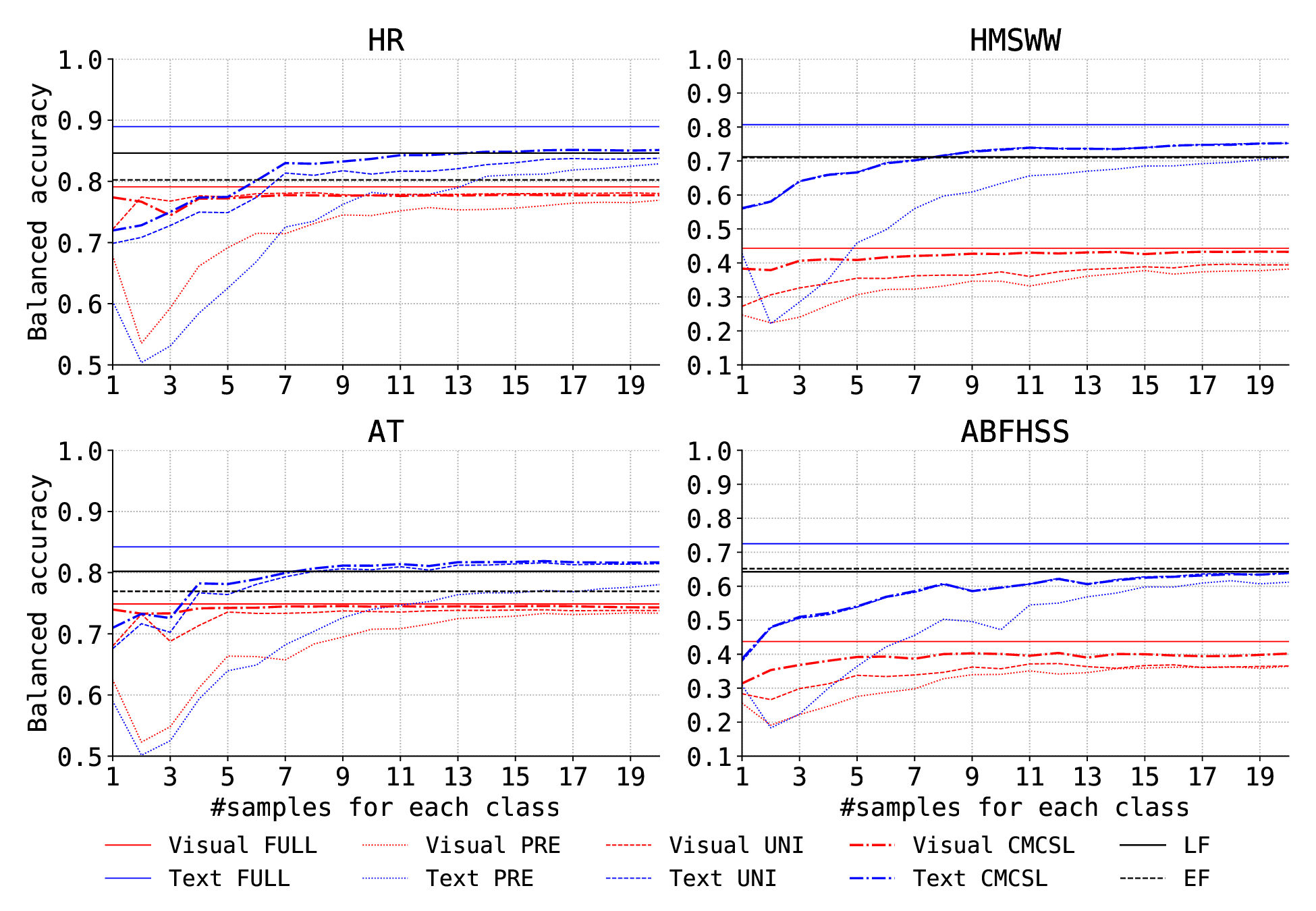}
    \caption{BAC in relation to the number of labeled samples for four example datasets.}
    \label{fig:ex3_all_datasets}
\end{figure}

Tables \ref{tab:ex3_binary} and \ref{tab:ex3_multi} present the results of comparing the classifier learned on \textsc{cmcsl} pseudo-labels with reference methods, averaged in terms of $b_{class}$. As can be seen, for both binary and multiclass problems, \textsc{cmcsl} is statistically significantly the best according to the Wilcoxon global rank test. In the case of binary data, for individual datasets, \textsc{cmcsl} is statistically comparable to \textsc{uni}, but tends to achieve a higher \emph{balanced accuracy score} for both modalities. In the case of multiclass problems, \textsc{cmcsl} allows us to obtain a statistically significantly best \textsc{bac} for visual modality while simultaneously demonstrating the ability to improve the generalization ability of the text modality classifier. These results allow us to answer \textbf{RQ2}, showing that for \emph{Gaussian Na\"ive Bayes}, the exchange of information between modalities in the self-labeling process allows for achieving generalization ability superior to a model learned solely on pseudo-labels obtained within a single modality. It is worth noting that \textsc{cmcsl} allows improving the quality of classification of visual modality even in cases where the use of conventional \textsc{lf} or \textsc{ef} undermines the obtained result due to a large discrepancy in the generalization ability of individual modalities. Fig. \ref{fig:ex3_all_datasets} presents the complete results in terms of the number of labeled samples for selected datasets.

Table \ref{tab:ex3_other_short} presents in a shortened form the results of Experiment 3 when the other two classification algorithms, rejected at the stage of Experiment 1, were used. Based on the Wilcoxon statistical test results, we can answer \textbf{RQ3} and conclude that at this stage of work, both \textsc{lr} and \textsc{cart} do not allow for increased generalization ability due to the use of \textsc{cmcsl} pseudo-labels. This is probably due to the inherent properties of classification algorithms and is a limitation of the proposed \textsc{cmcsl} method that should be considered in further research.

\begin{table}[!htb]
    \centering
    \scriptsize
    \caption{Wilcoxon test ($\alpha=0.05$) results for \textsc{lr} and \textsc{cart}. 
    }
   \renewcommand{\arraystretch}{0.7}
\begin{tabularx}{0.99\columnwidth}{l|CCC}
\toprule   
Classifier & PRE$^1$   & UNI$^2$   & CMCSL$^3$   \\
\midrule
& \multicolumn{3}{c}{Binary datasets} \\
\midrule
\textsc{cart} &        1.150  & \textbf{2.600 }   & 2.250   \\    
&  ---    & 1  & 1   \\
LR &  \textbf{2.100}  & 1.950    & 1.950   \\
 & ---    & ---      & ---     \\
 \midrule
& \multicolumn{3}{c}{Multiclass datasets} \\
\midrule
\textsc{cart} & 1.250  & \textbf{2.675}    & 2.075   \\    
& ---    & 1      & 1   \\
LR & \textbf{2.200}  & 1.875    & 1.925   \\
 & ---    & ---      & ---     \\
\bottomrule
\end{tabularx}
    \label{tab:ex3_other_short}
\end{table}

\begin{figure}[!htb]
    \centering
    \includegraphics[width=\columnwidth]{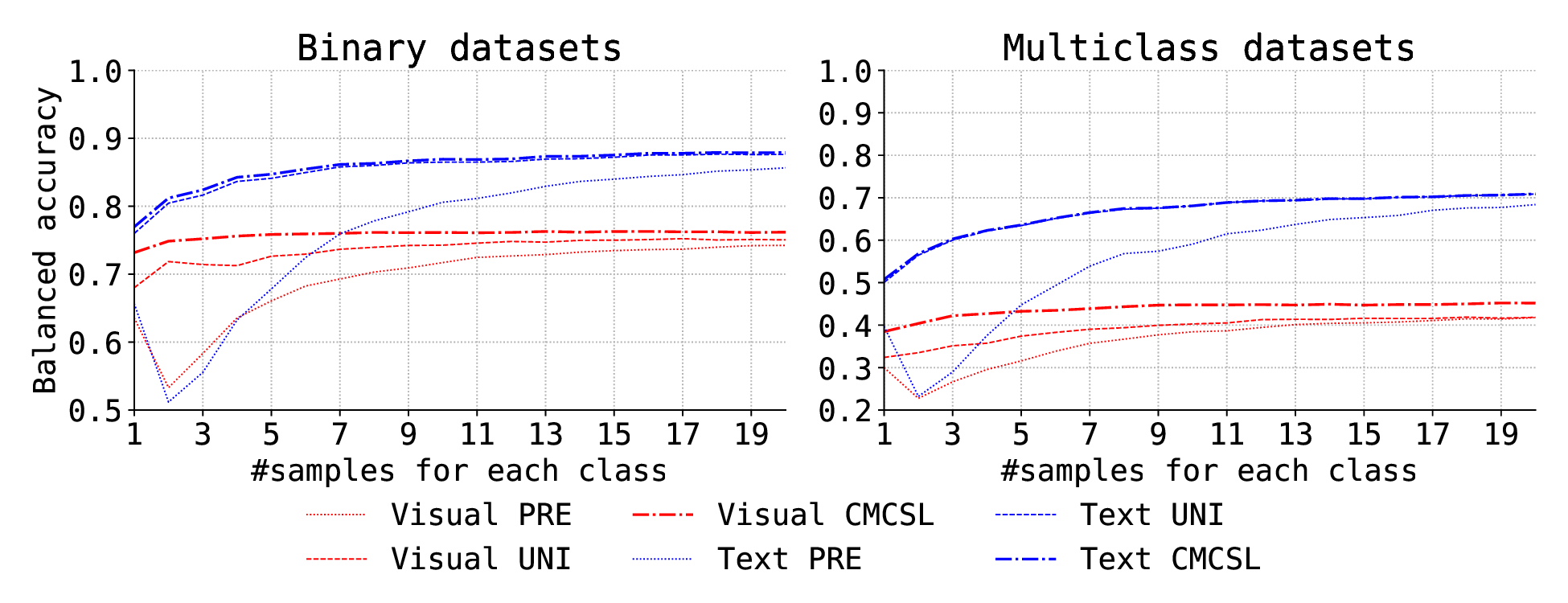}
    \caption{Visualization of BAC values versus number of labeled samples averaged for binary and multiclass datasets.}
    \label{fig:ex_times}
\end{figure}

In addition to the analysis for each dataset separately, averaging $b_{class}$ values, an analysis was also performed relative to the number of labeled samples for binary and multiclass datasets separately (averaging the results for datasets). The abbreviated results of the statistical analysis are shown in Tables. \ref{tab:ex3_binary_times} and \ref{tab:ex3_multi_times}, while Fig. \ref{fig:ex_times} shows the full results. It is worth noting that in the case of binary datasets, despite relatively small differences, CMCSL was found to be statistically significantly better not only in the global Wilcoxon singed-rank test, but also separately for individual numbers of labeled instances. For multiclass dataset, CMCSL performs statistically significantly better for visual modality both in terms of global ranks and for individual values of the $b_{class}$ parameter. For text modality, CMCSL is statistically comparable to UNI for individual labeled sample values, but achieves a statistically significant advantage in the Wilcoxon signed-rank test.

\begin{table}[!htb]
    \centering
    \scriptsize
    \caption{Comparison with reference methods for binary datasets in relation to the number of labeled samples for each class. 
    }
    \renewcommand{\arraystretch}{0.7}
\begin{tabularx}{0.99\columnwidth}{c|CCC|CCC}
\toprule
 &  & Visual &  & & Text & \\
\midrule
\makecell{Labeled\\samples}& PRE$^1$   & UNI$^2$   & CMCSL$^3$   & PRE$^1$   &   UNI$^2$ & CMCSL$^3$   \\
\midrule
 1             & 0.636 & 0.680 & \textbf{0.732}   & 0.657 & 0.760  & \textbf{0.770}   \\
               & ---   & 1     & 1       & ---   & 1     & all     \\
 3             & 0.583 & 0.714 & \textbf{0.752}   & 0.555 & 0.816 & \textbf{0.824}   \\
               & ---   & 1     & all     & ---   & 1     & all     \\
 5             & 0.661 & 0.726 & \textbf{0.758}   & 0.678 & 0.841 & \textbf{0.847}   \\
               & ---   & 1     & all     & ---   & 1     & all     \\
 7             & 0.693 & 0.737 & \textbf{0.760}   & 0.759 & 0.858 & \textbf{0.861}   \\
               & ---   & 1     & all     & ---   & 1     & all     \\
 9             & 0.709 & 0.742 & \textbf{0.761}   & 0.792 & 0.864 & \textbf{0.867}   \\
               & ---   & 1     & all     & ---   & 1     & all     \\
 11            & 0.725 & 0.746 & \textbf{0.761}   & 0.811 & 0.865 & \textbf{0.869}   \\
               & ---   & 1     & all     & ---   & 1     & all     \\
 13            & 0.729 & 0.747 & \textbf{0.763}   & 0.829 & 0.869 & \textbf{0.873}   \\
               & ---   & 1     & 1       & ---   & 1     & all     \\
 15            & 0.735 & 0.750 & \textbf{0.763}   & 0.840 & 0.872 & \textbf{0.875}   \\
               & ---   & 1     & all     & ---   & 1     & all     \\
 17            & 0.737 & 0.752 & \textbf{0.762}   & 0.847 & 0.875 & \textbf{0.878}   \\
               & ---   & 1     & all     & ---   & 1     & all     \\
 19            & 0.742 & 0.751 & \textbf{0.761}   & 0.854 & 0.876 & \textbf{0.879}   \\
               & ---   & 1     & 1       & ---   & 1     & all     \\

\midrule
\makecell{Average\\rank} & 1.000 & 2.000 & \textbf{3.000}   & 1.000 & 2.000     & \textbf{3.000}   \\
              & ---   & 1     & all    & ---   & 1     & all       \\
\bottomrule
\end{tabularx}
    \label{tab:ex3_binary_times}
\end{table}

\begin{table}[!htb]
    \centering
    \scriptsize
    \caption{Comparison with reference methods for multiclass datasets in relation to the number of labeled samples for each class. 
    }
    \renewcommand{\arraystretch}{0.7}
\begin{tabularx}{0.99\columnwidth}{c|CCC|CCC}
\toprule
 &  & Visual &  & & Text & \\
 \midrule
& PRE$^1$   & UNI$^2$   & CMCSL$^3$   & PRE$^1$   &   UNI$^2$ & CMCSL$^3$   \\
\midrule
\makecell{Average\\rank} & 1.000 & 2.000 & \textbf{3.000}   & 1.000 & 2.150  & \textbf{2.850}   \\
             & ---   & 1     & all    & ---   & 1     & all    \\
\bottomrule
\end{tabularx}
    \label{tab:ex3_multi_times}
\end{table}

\section{Conclusion}
The primary purpose of the work was to answer the question, does the exchange of propagated labels between modalities make it possible, within each modality, to build a classifier with a greater generalization ability than when no information exchange takes place? For this purpose, the \emph{Cross-Modality Clustering-based Self-Labeling} algorithm was proposed, which uses the Euclidean distance to propagate labels between modalities. As a result of the experiments carried out on the real-life MM-IMDb dataset subsets, it was confirmed that for \emph{Gaussian Na\"ive Bayes}, the use of pseudo-labels obtained thanks to \textsc{cmcsl} in the training process allows for statistically significantly better classification performance than in the absence of cross-modal information exchange. It is worth noting that while this gain is most pronounced for the less informative visual modality, \textsc{cmcsl} also shows the ability to statistically significantly improve classification performance in text modality and even in both modalities at the same time. The experiments also showed that each dataset should be treated individually to find the best approach for data preprocessing and transferring labels between modalities, and confirmed earlier observations made by Budnik et al. in \cite{budnik2014automatic}.
Future research may include: 
\textbf{(i)} adaptation of \textsc{cmcsl} for data stream semi-supervised classification tasks, \textbf{(ii)} employing the principle of cross-modal label propagation for the wide range of existing self-labeling methods, and \textbf{(iii)} examination of \textsc{cmcsl} properties when using modalities obtained in the data transformation process (e.g., using multi-dimensional encoding \cite{borisov2022deep}, sonification \cite{hermann2008taxonomy}, or sentence space \cite{lopez2017deep}.

\bibliographystyle{IEEEtran}
\bibliography{bibliography}

\end{document}